\documentclass[sigconf]{acmart}
\renewcommand\footnotetextcopyrightpermission[1]{}
\AtBeginDocument{%
  }

\usepackage[ruled,linesnumbered]{algorithm2e}
\usepackage{multirow}
\usepackage{multicol}

\setcopyright{acmcopyright}
\copyrightyear{2023}
\acmYear{2023}
\acmDOI{XXXXXXX.XXXXXXX}

\acmConference[MM '23]{Proceedings of the 31st ACM International Conference on Multimedia}{October 29--November 02, 2023}{Ottawa, Canada}

\acmPrice{15.00}
\acmISBN{978-1-4503-XXXX-X/18/06}

\begin{document}

\title{CSSL-RHA: Contrastive Self-Supervised Learning for Robust Handwriting Authentication}


\author{Jingyao Wang}
\affiliation{%
  \institution{Institute of Software, Chinese Academy of Sciences}
  \city{Beijing}
  \country{China}}
\email{jingyao_wang0728@163.com}

\author{Luntian Mou*}
\affiliation{%
  \institution{Beijing University of Technology}
  \city{Beijing}
  \country{China}}
\email{ltmou@pku.edu.cn}

\author{Changwen Zheng*}
\affiliation{%
  \institution{Institute of Software, Chinese Academy of Sciences}
  \city{Beijing}
  \country{China}}
\email{changwen@iscas.ac.cn}

\author{Wen Gao}
\affiliation{%
  \institution{Peking University}
  \city{Beijing}
  \country{China}}
\email{wgao@pku.edu.cn}






\renewcommand{\shortauthors}{Author 1 and Author 2, et al.}


\begin{abstract}
  Handwriting authentication is a valuable tool used in various fields, such as fraud prevention and cultural heritage protection. However, it remains a challenging task due to the complex features, severe damage, and lack of supervision. In this paper, we propose a novel Contrastive Self-Supervised Learning framework for Robust Handwriting Authentication (CSSL-RHA) to address these issues. It can dynamically learn complex yet important features and accurately predict writer identities. Specifically, to remove the negative effects of imperfections and redundancy, we design an information-theoretic filter for pre-processing and propose a novel adaptive matching scheme to represent images as patches of local regions dominated by more important features. Through online optimization at inference time, the most informative patch embeddings are identified as the "\textit{most important}" elements. Furthermore, we employ contrastive self-supervised training with a momentum-based paradigm to learn more general statistical structures of handwritten data without supervision. We conduct extensive experiments on five benchmark datasets and our manually annotated dataset EN-HA, which demonstrate the superiority of our CSSL-RHA compared to baselines. Additionally, we show that our proposed model can still effectively achieve authentication even under abnormal circumstances, such as data falsification and corruption.

\end{abstract}





\maketitle

\section{INTRODUCTION}
\label{introduction}

Handwriting refers to the act of writing using a specific instrument (e.g., a pen or pencil). Since each person's handwriting is unique, it is considered an important characteristic that finds application in various fields such as identity verification, e-security, and e-health \cite{cpalka2016new, xu2017challenge, santangelo2016comprehensive}. For instance, doctors utilize changes in handwriting as a diagnostic sign for Neurodegenerative Diseases (NDs) \cite{de2019handwriting}, while forensic experts use handwriting clues to identify suspects \cite{morris2020forensic}. Therefore, the utilization of handwriting characteristics has emerged as a prominent interdisciplinary topic with many excellent studies exploiting it in a supervised manner \cite{sharipov2019analysis, faundez2020handwriting}. However, they usually demand annotated handwritting data which requires a lot of workforce.

Picture a scenario where we create archives using written materials such as books, papers, or even ancient bamboo slips. These archives consist of handwriting data from a large number of individuals uploaded in real time to collect a significant amount of unlabeled signals. However, this process requires an exceedingly intricate and time-consuming labeling process. Furthermore, due to the susceptibility of handwriting data to noise and potential corruption, many of the existing supervised structures that are designed for labeled and unblemished data may not perform well. To overcome these barriers in practical applications, self-supervised learning (SSL) has garnered increasing attention in recent years \cite{jaiswal2020survey, misra2020self}. This data-efficient paradigm focuses on decoding representations with generalization capabilities.

Several studies have delved into self-supervised learning to acquire representations of handwriting features, which can overcome the scarcity of labeled data \cite{lastilla2022self, chattopadhyay2022surds}. Gidaris et al. \cite{gidaris2020learning} proposed a self-supervised method that relies on spatially dense image descriptions to encode discrete visual concepts and achieve exceptional performance. However, in cases where the data is corrupted or encounters significant disturbances, obtaining representations of the original data from the corrupt inputs can result in significant ambiguities. This issue is particularly challenging in handwriting authentication, as it involves a vast number of fake samples.

Based on our observations mentioned above and the findings of previous works, our aim is to address two main criteria: 1) developing a comprehensive understanding of the structures and statistical features of handwriting data, which can generalize towards new classes; and 2) providing the capability to interpret provided instances even in the presence of corrupted or falsified images.

\begin{figure}
  \includegraphics[width=0.5\textwidth]{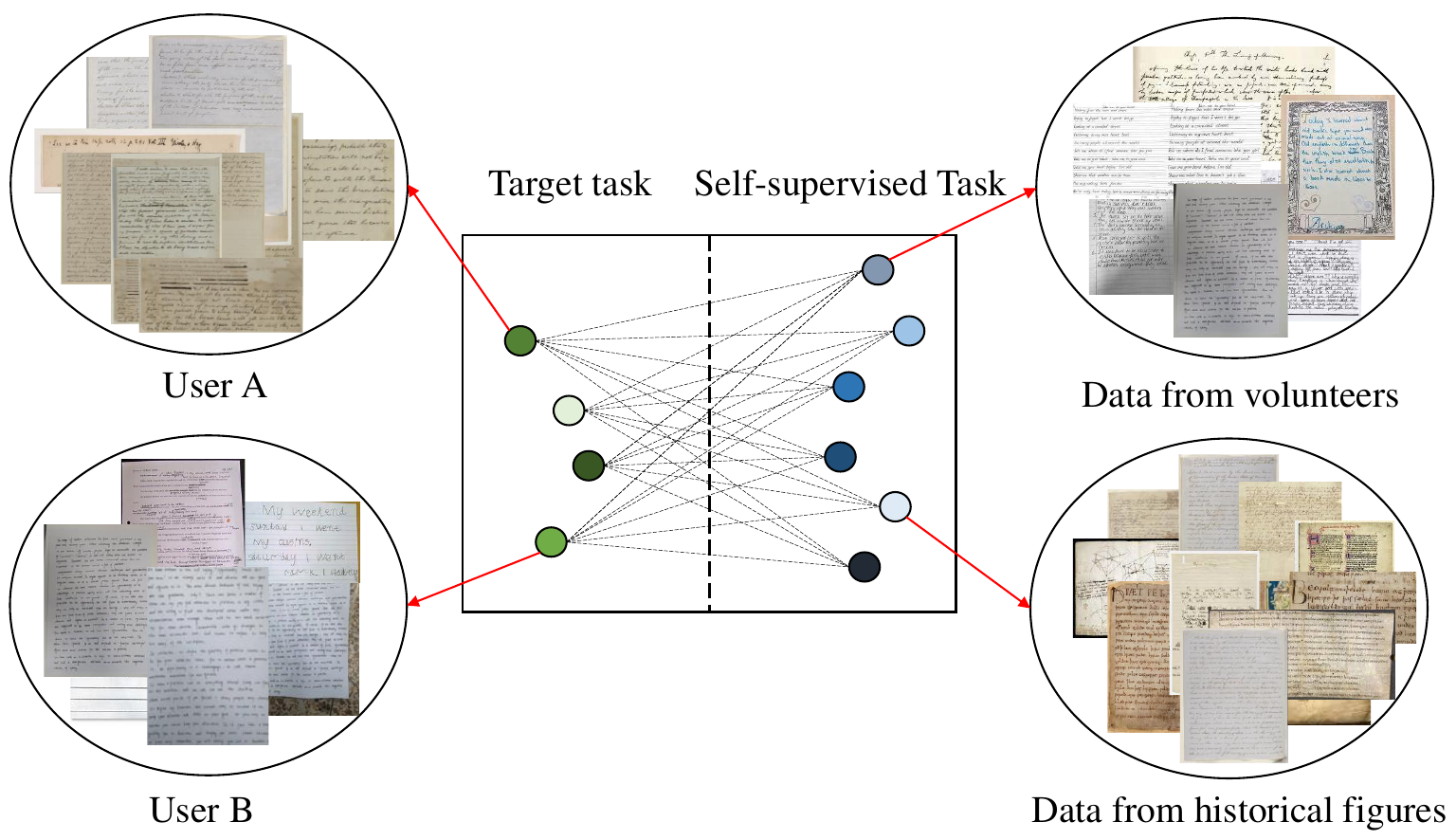}
    \caption{The general framework of CSSL-RHA. The green nodes represent the embedding of categories in the target task, and the blue nodes represent the embedding of domains in the self-supervised task. The dotted lines denote the connection between them established through the pre-training of CSSL-RHA.}
    \label{fig:teaser}
\end{figure}

To address these challenges, we propose a novel Contrastive Self-Supervised Learning framework for Robust Handwriting Authentication (CSSL-RHA). To eliminate the negative impact of defects and redundancies, we design an information-theoretic filter for pre-processing, and decompose images into patches representing local regions that are likely to be dominated by more important features through a novel adaptive matching scheme. Since fine-grained annotations may be lacking, we employ contrastive self-supervised training with a momentum-based paradigm that to establish semantic patch correspondences efficiently. Our proposed CSSL-RHA method effectively extracts and utilizes important information from increasing amounts of unlabeled and damaged data. The general framework of our proposed CSSL-RA is depicted in Figure \ref{fig:teaser}.

Our contributions can be summarized as follows:
\vspace{-\topsep}
\begin{itemize}
\item We propose a \textbf{C}ontrastive \textbf{S}elf-\textbf{S}upervised \textbf{L}earning framework for \textbf{R}obust \textbf{H}andwriting \textbf{A}uthentication (CSSL-RHA), which employs a momentum-based paradigm for pretext tasks and an novel adaptive matching scheme as an accelerator to comprehend and interpret data.
\item We introduce a dedicated dataset for Handwriting Authentication, called EN-HA, consisting of 800 English manuscripts collected from 20 famous historical figures and 20 volunteers. The dataset simulates real-world situations, including data damage and falsification. 
\item Extensive experiments demonstrate the superiority of the proposed CSSL-RHA approach on five benchmark datasets and our EN-HA, and additional experiments confirm the effectiveness of our method from various perspectives.
\end{itemize}

\begin{figure*}
  \includegraphics[width=\textwidth]{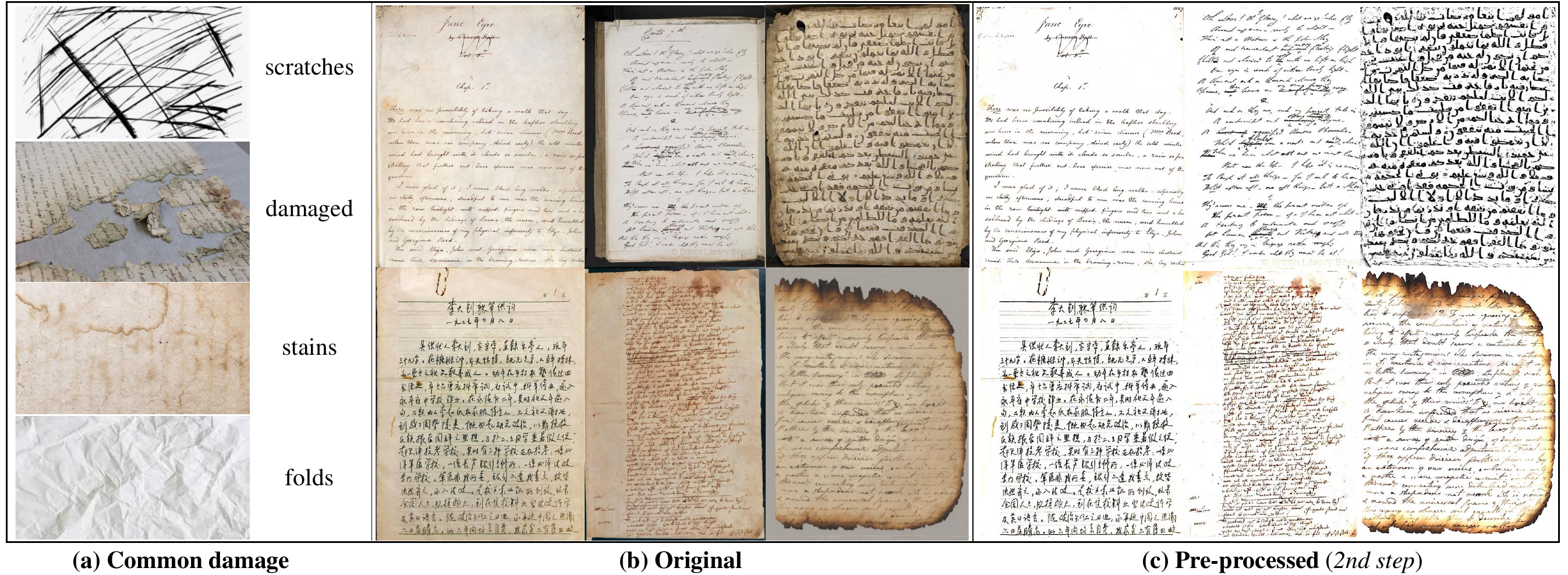}
    \caption{Handwriting defects and pre-processing results. (a) the common defects and damages of the collected manuscripts; (b) samples of the collected handwriting data; (c) samples after pre-processing with 2 steps.}
\label{fig:defect}
\end{figure*}

\section{RELATED WORK}
\label{related work}

\subsection{Handwriting Authentication}
\label{2-Handwriting Authentication}

Unlike physiological characteristics, handwriting is a behavioral characteristic, meaning that no two individuals with mature handwriting are identical, and one individual cannot produce another's handwriting exactly \cite{rehman2019writer}. Handwriting has gained increasing attention and found widespread applications in various fields such as forensic identification \cite{morris2020forensic}, medical diagnosis \cite{de2019handwriting}, information security \cite{faundez2020handwriting}, and document forgery detection \cite{adak2019empirical}. However, due to its complexity and large variability, handwriting authentication remains a challenging task.

To cope with the growing demand for handling big data, traditional manual inspection of handwriting style is inefficient \cite{khan2019knn, gupta2020improved}. As a result, research has shifted towards machine-based methods to bridge the gap between current needs and actual applications. Davis et al. \cite{davis2020text} developed a generator network trained with GAN and autoencoder techniques to learn handwriting style. Zhang et al. \cite{zhang2016end} proposed an end-to-end framework for online text-independent writer identification that leverage a recurrent neural network (RNN). Lian et al. \cite{lian2016automatic} introduced a system that can automatically extract and characterize personal handwriting styles in a font library. Zhu et al. \cite{zhu2016offline} presented an offline writer identification system using sparse auto-encoder based codebook and non-text-segmentation feature extraction methods. Although these supervised approaches have been successful in enhancing the efficiency of handwriting authentication, it is difficult for them that require labeled and sound data to obtain similar effects due to the defects in daily applications (see Figure \ref{fig:defect} (a)).

\subsection{Self-supervised Learning}
\label{2-Self-supervised}
Self-supervised learning has played a crucial role in advancing technology during the big data era, offering robust representation by transferring knowledge from pretext tasks without requiring extra annotations. Notably, recent studies in vision-based self-supervised tasks can be broadly categorized into two types: contrastive learning and generative learning.

\noindent\textbf{Contrastive learning} is a technique that involves learning the shared characteristics between different data classes while distinguishing one class from another by contrasting examples. This technique can significantly enhance performance in various visual tasks. For instance, SimCLR \cite{chen2020simple} is a simplified contrastive self-supervised learning approach for visual representations, which achieves outstanding results in computer vision without requiring specialized architectures or a memory bank. For handwriting data, Lin et al. \cite{lin2022cclsl} proposed an encoder-decoder method using contrastive learning to effectively learn semantic-invariant features between printed and handwritten characters. Viana et al. \cite{viana2022contrastive} explained how deep contrastive learning can lead to an improved dissimilarity space for learning handwritten signature feature representations. Zhang et al. \cite{zhang2022cmt} extended SimCLR to a character-level pretext task, known as Character Movement Task, showing the significant advantages of self-supervision via pretext tasks in dealing with complex handwriting features. Despite the underrepresentation of self-supervised learning in authentication, these works demonstrate its potential in effectively addressing the intricate nature of handwriting data.

\noindent\textbf{Generative learning} involves the reconstruction of inputs from original or corrupted inputs, and the creation of representation distributions at a point-wise level, such as pixels in images and nodes in graphs. Key methods within this paradigm include auto-regressive (AR) models \cite{matero2020autoregressive, you2018graphrnn}, flow-based models \cite{dohi2021flow, kingma2018glow}, auto-encoding (AE) models \cite{kipf2016variational, sabokrou2019self}, and hybrid generative models \cite{dai2019transformer, khajenezhad2020masked}. For handwriting data, considering the properties of this paradigm, it can work in terms of data augmentation for auxiliary training when data is insufficient or damaged. He et al. \cite{he2022masked} introduced masked autoencoders (MAE) as a scalable self-supervised learning approach for computer vision aim to reconstruct missing patches in images. However, although generative learning can mitigate the negative effects of lacking labels or damaged images, it is not suitable for handwriting authentication whose primary focus is learning complex features without supervision.

\section{METHODOLOGY}
\label{Methodology}
We begin this section by briefly introducing the formulation in this work. Our proposed method, CSSL-RHA (Contrastive Self-Supervised Learning for Robust Handwriting Authentication), is then presented in an overview (see Figure \ref{fig:CSSL-RHA}), followed by a detailed elaboration on its four stages.

\subsection{Formulation}
\label{Formulation}
We adhere to the standard self-supervised learning pipeline, which involves self-supervised pre-training followed by task-specific fine-tuning. During the pre-training phase, we combine the unlabeled training data from all subjects denoted as $I=\left \{ \mathcal{I} _{1},...,\mathcal{I} _{N} \right \}\in \mathbb{R}^{H\times W\times C} $, where $N$ is the number of the subjects, $H\times W$ denotes the resolution of the images, and $C$ represents the number of channels. The data is pre-processed to eliminate any defects and represented as $X=\left \{ x_{1},...,x_{N} \right \} \in \mathbb{R}^{H\times W\times C}$. The calibration data and label are represented as $I_{s}^{c}$ and $Y_{s}^{c}$, while the test data and label are denoted as $I_{t}^{c}$ and $Y_{t}^{c}$. Our proposed CSSL-RHA is defined by the parameters $\theta$, and the pre-trained general encoder is denoted by $F_{\theta}$, which generates the tokens set $Z_{x}$. The contrastive learning branch comprises the patch head, projection head, and prediction head, represented by $P^{pat}(\cdot)$, $P^{pro}(\cdot)$, and $P^{pre}(\cdot)$, respectively. To facilitate contrastive learning, a momentum branch is adopted and composed of the same encoder $F_{\theta}$, a patch head $K^{pat}(\cdot)$, and a projection head $K^{pro}(\cdot)$. During the task-specific fine-tuning phase, the model consists of the encoder $F_{\theta}$, which is trained during the self-supervised pre-training phase, and a task-specific decoder.

\subsection{Overview}
\label{Overview}
\begin{figure*}
  \includegraphics[width=\textwidth]{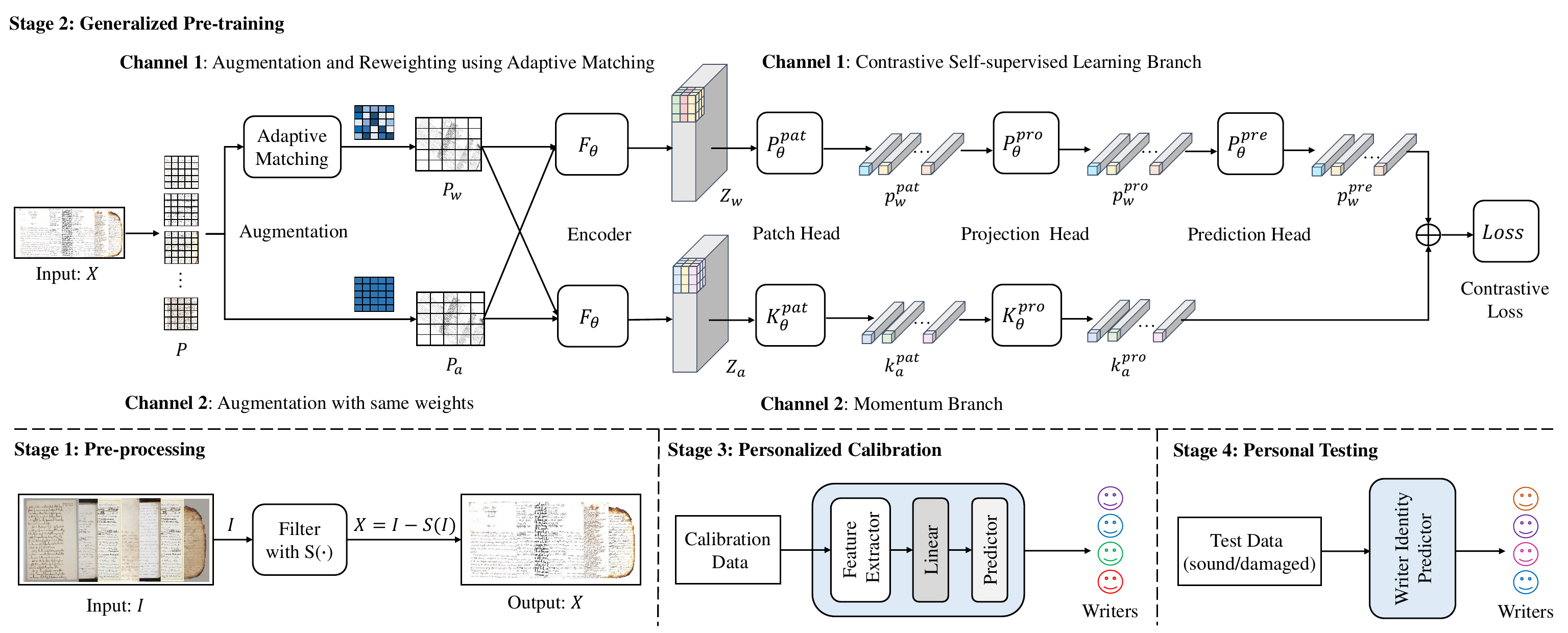}
    \caption{The overall process of our proposed CSSL-RHA with four stages based on a CNN-Transformer hybrid structure.}
    \label{fig:CSSL-RHA}
\end{figure*}

We propose a Contrastive Self-Supervised Learning method for Robust Handwriting Authentication (CSSL-RHA), which utilizes an novel adaptive matching scheme with a momentum-based paradigm, as illustrated in Figure \ref{fig:CSSL-RHA}. The entire model consists of four stages: pre-processing, generalized pre-training, personalized calibration, and personal testing. In the pre-processing stage, the unlabeled handwriting data $I$ from all subjects undergo processing to mitigate any negative impacts caused by paper damage, stains, or other defects, obtaining $X$. In the pre-training stage, the pre-pocessed images $X$ are augmented, reweighted and compared with each other to learn the general information extracted by $F_{\theta}$, which is shared by all subjects. In the personalized calibration stage, only a small amount of labeled data $I_{s}^{c}$ and $Y_{s}^{c}$ from a specific subject $s$ are used to calibrate the personal handwriting predictor from the pre-trained generalized encoder $F_{\theta}$. In the testing stage, both clear and damaged handwriting data $I_{t}^{c}$ can be decoded to recognize the writer identities.

\begin{figure}
  \includegraphics[width=0.5\textwidth]{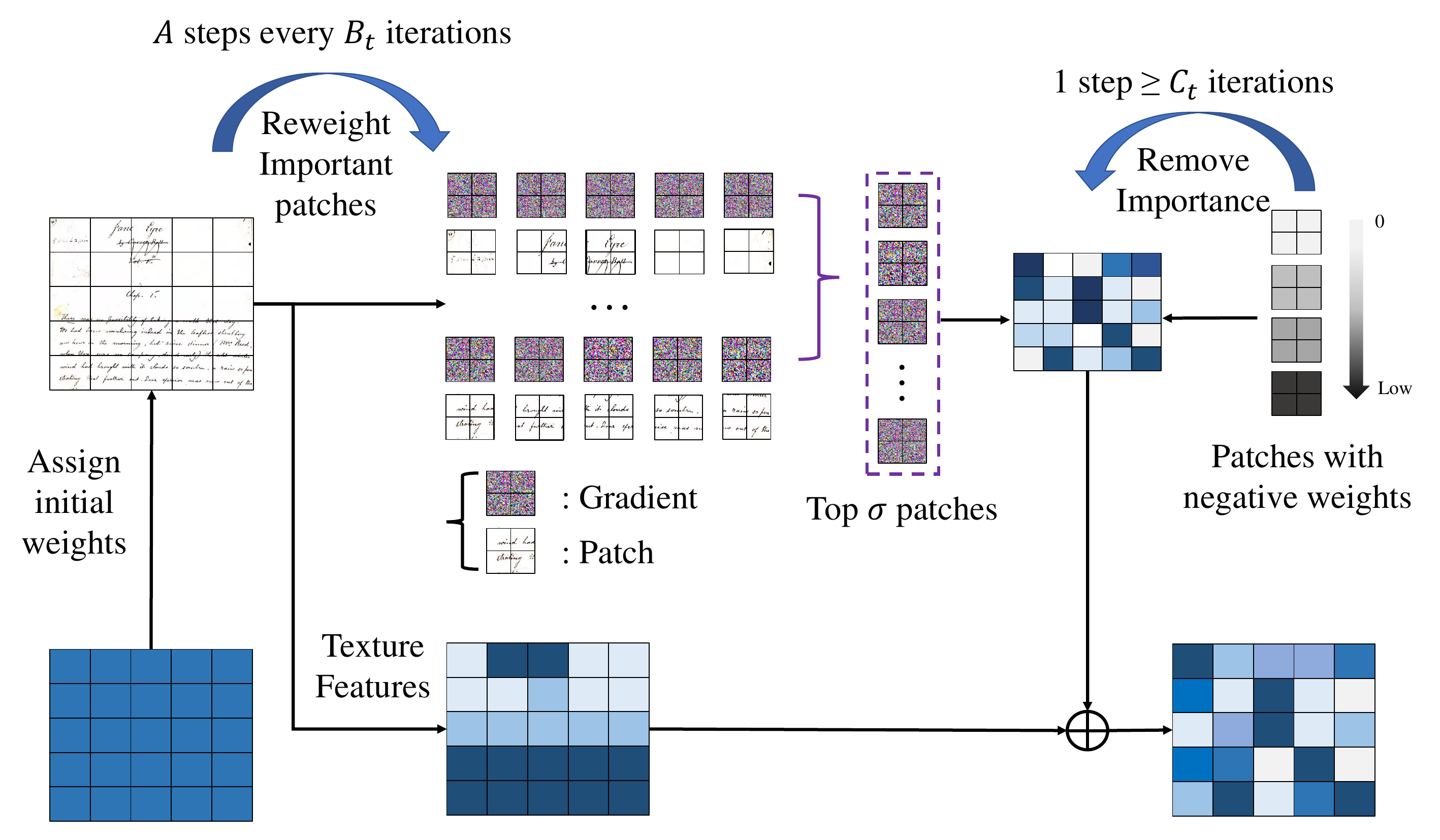}
    \caption{Learning importance through adaptive matching. The most helpful patches for the task at hand are determined via optimization at inference time by reweighting all patches based on their contribution towards a correct classification result.}
    \label{fig:Learning patch importance}
\end{figure}

\subsection{Pre-processing}
\label{Pre-processing}
Handwriting classification can be adversely affected by various factors, such as improper storage of paper or external intervention, which can significantly reduce its effectiveness. Common text image defects that affect handwriting quality include scratches, damage, stains, and folds (as shown in Figure \ref{fig:defect} (a)). While authentic works of famous authors are valuable objects, they still have image defects that can impact the visual quality of handwriting manuscripts. For instance, connecting strokes under stain masking can be mistakenly judged as high density handwriting dots.

To improve the visual quality of manuscripts after pre-processing, we design an information-theoretic filter. The filter addresses disturbances caused by handwriting condition and text preservation in daily life. Figure \ref{fig:defect} (b) shows some samples of authentic works of famous authors collected in our study. We start by obtaining the input handwriting data $I=\left \{ \mathcal{I} _{i} \right \}_{i=1}^{N}$. Next, we use a texture feature detection operator $S(\cdot)$ to get the noise residual images $I^{'}=\left \{ \mathcal{I} _{i}^{'} \right \}_{i=1}^{N}$ and obtain denoised data $X$ by:
\begin{equation}
\begin{aligned}
x_{i}=\mathcal{I} _{i}-\mathcal{I} _{i}^{'}=\mathcal{I} _{i}-S(\mathcal{I} _{i}) \\
s.t.\quad S(\cdot)=E_{B}(\cdot)/E_{N}(\cdot)
\end{aligned}
\end{equation}
where $E_{N}$ and $E_{B}$ are the average energy and the local energy of noise respectively calculated as follows:
\begin{equation}
\begin{aligned}
E_{N}(\mathcal{I} _{i})=\frac{1}{H\times W} \sum_{t=1}^{T} \left \| FFT_{t}^{2}
(\mathbb{I}_i )\right \| _{2}^{2} \\
E_{B}(\mathcal{I} _{i})=w_t  \sum_{t=1}^{T} \left \| FFT_{t}^{2}
(\mathbb{I}_i )\right \| _{2}^{2} \\
s.t. \quad \mathbb{I}_i=\sum_{j=1}^{H\times W}(\lambda R(\mathcal{I}_i^j),D(\mathcal{I}_i^j,I) ) 
\end{aligned}
\end{equation}
Here, $FFT^{2}$ denotes the two dimensional Fast Fourier Transform; $\left \| \cdot \right \|_{2} $ represents the 2-norm, $\left \| \cdot \right \| _{2}^{2}$ denotes the energy at a point in the Fourier transform frequency domain, and $w_{t}$ is a normalized radially symmetric window function. For constraint $\mathbb{I}_i$, $R(\cdot)$ is the regularization term used to maintain the continuity and smoothness, $D(\cdot,\cdot)$ is the data term consisting of reconstruction error and similarity constraint, and $\lambda$ is the regularization parameter.

After obtaining $E_{N}(\cdot)$ and $E_{B}(\cdot)$, we use the operator $S(\cdot)$ to re-identify the filtered residual image and extract the misclassified texture detail information. Finally, we compensate for the detected detail information into the filtered image to obtain the final denoised image $X$. The visual quality improvement of manuscripts after pre-processing is shown in Figure \ref{fig:defect} (c).

\subsection{Generalized Pre-training}
\label{Generalized Pre-training}

During this stage, we aim to pre-train the generalized feature encoder, $F_{\theta}$, which is designed to learn from unlabeled handwriting data from all subjects. The primary goal is to improve the recognition of writer identities for a specific subject, even when dealing with few and damaged data. To achieve this, we propose a momentum-based contrastive self-supervised learning approach with a novel adaptive matching scheme to learn general representations. Our pre-training model is designed based on a hybrid CNN-ViT structure, which takes into consideration the characteristics of handwriting. To the best of our knowledge, this is the first successful application of self-supervised learning in the field of multi-person handwriting authentication.

\subsubsection{Authentication through CSSL-RHA}
\
\newline
As illustrated in Figure \ref{fig:CSSL-RHA}, we augment each denoised images $X=\left \{ x_{1},...,x_{N} \right \} \in \mathbb{R}^{H\times W\times C}$, and split each image into a sequence of $M=H\times W/\mathcal{P}^2 $ patches, denoted by $P=\left \{ p_i \right \}_{i=1}^M$, with each patch $p_i\in \mathbb{R}^{\mathcal{P}^2\times C}$. Then, we split them into two channels: the first channel uses our proposed adaptive matching scheme to identify the most significant component, which results in reweighted patches denoted by $P_{w}$; and 2) the other channel assigns the same weights to the patches of the augmented images as input and is denoted by $P_{a}$.

We then flatten and input all patches from $P_{w}$ and $P_{a}$ to the Transformer encoder $F_{\theta}$, obtaining the set of $P_{w}$ tokens denoted by $Z_w=F_{\theta}(P_{w})$ with $Z_w=\left \{ z_{w}^{n}|n=1,...,N \right \} $, and the set of $P_{a}$ tokens denoted by $Z_a=F_{\theta}(P_{a})$ with $Z_a=\left \{ z_{a}^{n}|n=1,...,N \right \} $. Each element in $Z_w$ and $Z_a$ is a vector of length $L$ and belongs to $\mathbb{R}^D$. We select two types of Transformers for the feature encoder: Vision Transformers, such as ViT \cite{dosovitskiy2020image}, and hierarchical Transformers, such as Swin \cite{liu2021swin}. ViT satisfies $L=M$, whereas Swin generally produces a reduced number of tokens $L<M$ due to internal merging strategies.

After obtaining patch embeddings for both $Z_{w}$ and $Z_{a}$, we again split them into two channels: the contrastive self-supervised learning branch and the momentum branch \cite{he2020momentum}. In the first channel, we sequentially process the patch embeddings $Z_{w}$ through $P_{\theta}^{pat}$, $P_{\theta}^{pro}$, and $P_{\theta}^{pre}$. The patch head $P_{\theta}^{pat}$ projects the encoded feature maps $Z_{w}$ into a single vector $p_{w}^{pat}$, which serves as the atomic element in a contrastive loss. The projection head $P_{\theta}^{pro}$ and prediction head $P_{\theta}^{pre}$ consist of a 3-layer MLP and a 2-layer MLP, respectively, to further process $p_{w}^{pat}$, where each layer includes a fully-connected layer and a normalization layer. In the second channel, we use $K_{\theta}^{pat}$ and $K_{\theta}^{pro}$, which have the same architecture as the contrastive learning branch, to process $Z_{a}$. They are parameterized by an exponentially moving average (EMA) of their parameters, respectively.

After obtaining patch embeddings $p_{w}^{pre}$ for the first channel and $k_{a}^{pre}$ for the other channel for the same batch, we establish semantic correspondences and broadcast them to compute the loss. It is important to note that the addition of our reweighting logits corresponds to multiplicative reweighting in probability space. In this study, we consider a form of a contrastive loss function, called InfoNCE \cite{oord2018representation}, for handwriting authentication:
\begin{equation}
    \mathcal{L}=-log\frac{exp(p\cdot k_{+}/\tau )}{ {\textstyle \sum_{i=1}^{N}exp(p\cdot k_{i}/\tau)} }  
\end{equation}
where $\tau$ is a temperature hyper-parameter. The summation is taken over one positive sample and $N-1$ negative samples. Essentially, this loss is the logarithmic loss of a $N$-way softmax classifier that attempts to classify $p$ as $k_{+}$. The objective of the self-supervised learning pre-training stage is to obtain general information from the process described above.

\subsubsection{Learning importance through adaptive matching}
\
\newline
In the pre-training, we propose an adaptive matching scheme to learn the importance of each individual patch token form all samples of $X$ via online optimization at inference time, as illustrated in Figure \ref{fig:Learning patch importance}. 

Due to the complexity and diversity of handwriting in manuscripts, it is difficult to locate key areas using simple texture judgments. To address this issue, we propose an adaptive method where we add $\sigma$ patches with a weight boost of $\alpha$ to the image with $A$ steps based on the current gradient in every $B_t$ iterations. This approach allows us to find new key-patches that are most suitable for the current iteration.

Although the augmented samples mostly retain the same properties \cite{cubuk2019autoaugment}, the key patches we compute may have different importance. While setting a consistent weight matrix for homologous augmented samples in a batch can improve computational efficiency, it may also lead to performance degradation. Adjusting the patch weight can make full use of the data. In our work, we remove patches with changes smaller than the mean change divided by three until we reach a maximum number of iterations $T_{MAX}$ (the minimum number of bounding boxes exceeding the threshold is at least $C_t$ times).

\subsection{Personalized Calibration and Test}
\label{Personalized Calibration and Test}
During the calibration stage, the data for a particular subject consist of a few labeled samples from each writer identity in the original training dataset. These samples and labels are represented as $I_{s}^{c}$ and $Y_{s}^{c}$. Since the handwriting data has no inherent order relationship, it is reasonable to randomly select non-repetitive training data as calibration data. To obtain a personalized calibrated predictor, we fine-tune $F_{\theta}$ followed by a linear layer to predict the writer identity.

During the test stage, our model is capable of accepting both sound and damaged handwriting data. To verify the effectiveness of the personalized model, we use the test set of subject $s$ from the original test dataset, represented as $I_{t}^{c}$ and $Y_{t}^{c}$. To ensure the reliability of the test results, we apply the same weight calculation and enhancement method as in the pre-training stage.

 

\section{Experiments}
\label{Experiments}

\subsection{Datasets}
\label{Datasets}
Our CSSL-RHA is evaluated on five benchmark datasets covering English language and mixed language, and one self-collected dataset EN-HA (with falsified and damaged data).

\textbf{IAM} \cite{marti1999full} is extensively used for handwriting authentication, containing 13,353 labeled text lines of variable content written in English, with approximately 14 text lines per writer. It includes 1,539 forms produced by 657 different writers, providing detailed information such as writer identity and ground truth text.

\textbf{CEDAR} \cite{srihari2002individuality} is a database with lines of English text, handwritten on a writing tablet by around 200 writers. The database includes 105,573 words, and we randomly stitch together text images of the same author in this work.

\textbf{CVL} \cite{kleber2013cvl} is a mixed language dataset consisting of both English and German handwritten text of 310 different writers. The database includes 7 different handwritten texts, each of which has an RGB color image and a cropped version. A total of 310 writers participated in the dataset, with 27 of them writing 7 texts and 283 writers writing 5 texts.

\textbf{QUWI} \cite{al2012quwi} includes 1,068 digitized handwriting pages gathered from 1,017 writers in both Arabic and English scripts. Each individual was asked to write 4 pages, and only the data from the first and third pages are used in our study for the text-independent handwriting authentication task.

\textbf{ICDAR2013} \cite{hassaine2013icdar} is collected from 475 writers on 4 handwritten documents in both English and Arabic script as similar to QUWI. In this study, we apply both languages, where the first and second pages have Arabic handwritten text while the third and fourth contain English samples.

\textbf{EN-HA} is a label-free English handwriting dataset for handwriting authentication consisting of 800 English manuscripts collected from 20 famous historical figures and 20 volunteers. The manuscripts of famous authors are collected from open source data from the British Museum, Vatican Museum, etc. This dataset simulates real-world situations, including 90\% images (with average noise area about 10\%), and 10\% falsified images.

\subsection{Implementation Details}
\label{Implementation Details}
For the feature encoder, we select two types of Transformers: vision Transformers like ViT \cite{dosovitskiy2020image}, and hierarchical Transformers like Swin \cite{liu2021swin} in order to compare with various state-of-the-art writer identity recognizers with different model sizes. The ViT is adopted as the default backbone with the embedding size of 512. For adaptive matching, the step size of reweighting the patches is set to 3 ($A=3$), $\sigma$ is set to 10, $B_{t}$ and $C_{t}$ are set to 10 and 20, respectively. For optimizer, we use Adam \cite{kingma2014adam} to train models, where $\beta_{1}$= 0.5 and $\beta_{1}$ = 0.6 in the optimizer. The training hyperparameters are: the base learning rate as 1e-4, the batch size as 1,024, and the weight decay of 0.05. Considering the nature and texture characteristics of handwriting data, we apply data augmentations including random Gaussian blurring, random mixup and random  horizontal flip across clips. All experiments are conducted with 8 NVIDIA V100 (32GB RAM) GPUs.

\begin{table*}
  \centering
  \caption{Comparison with baseline models on five benchmark datasets and our EN-HA dataset. The values in the table represent the Accuracy(\%), and the optimal values will be marked in bold.}
  \label{tab:comparison}
  \begin{tabular}{l|cc|cc|cc|cc|cc|cc}
    \toprule
    \multirow{2}{*}{\textbf{Methods}} & \multicolumn{2}{c|}{\textbf{IAM}} & \multicolumn{2}{c|}{\textbf{CEDAR}} & \multicolumn{2}{c|}{\textbf{CVL}} & \multicolumn{2}{c|}{\textbf{QUWI}} & \multicolumn{2}{c|}{\textbf{ICDAR2023}} & \multicolumn{2}{c}{\textbf{EN-HA}}\\
    \cline{2-13}
    & \textbf{Top 1} & \textbf{Top 5} & \textbf{Top 1} & \textbf{Top 5} & \textbf{Top 1} & \textbf{Top 5} & \textbf{Top 1} & \textbf{Top 5} & \textbf{Top 1} & \textbf{Top 5} & \textbf{Top 1} & \textbf{Top 5} \\
    \midrule
    \textbf{SimCLR} \cite{chen2020simple} & 61.412 & 80.515 & 75.832 & 82.012 & 62.301 & 85.298 & 49.785 & 56.353 & 51.936 & 69.501 & 45.893 & 66.289 \\
    \textbf{BYOL} \cite{grill2020bootstrap} & 53.238 & 79.789 & 71.520 & 79.233 & 49.872 & 82.011 & 50.542 & 68.487 & 55.293 & 71.220 & 44.732 & 71.458 \\
    \textbf{Barlow Twins} \cite{zbontar2021barlow} & 49.947 & 86.289 & 71.389 & 79.278 & 62.332 & 80.299 & 42.891 & 57.777 & 50.829 & 68.839 & 48.203 & 68.128 \\
    \textbf{MOCO} \cite{he2020momentum} & 64.852 & 82.545 & 69.298 & 80.122 & 58.825 & 71.513 & 53.044 & 61.825 & 49.063 & 74.202 & 56.256 & 77.197 \\
    \midrule
    \textbf{NN-LBP} \cite{chahi2019effective} & 18.512 & 31.293 & 24.355 & 39.053 & 13.523 & 28.238 & 9.328 & 17.938 & 19.544 & 35.205 & 10.083 & 21.279 \\
    \textbf{NN-LPQ} \cite{chahi2019effective} & 18.148 & 32.932 & 25.534 & 37.562 & 14.200 & 30.856 & 10.254 & 17.652 & 17.025 & 34.545 & 12.877 & 22.830 \\
    \textbf{NN-LTP} \cite{chahi2019effective} & 17.843 & 29.842 & 24.378 & 37.234 & 14.784 & 30.239 & 9.010 & 16.382 & 21.019 & 38.231 & 11.793 & 24.873 \\
    \textbf{CoHinge} \cite{he2017beyond} & 19.622  & 35.215 & 40.420 & 51.527 & 18.164 & 34.055 & 15.058 & 22.024 & 22.027 & 44.789 & 14.109 & 26.724 \\
    \textbf{QuadHinge} \cite{he2017beyond} & 20.984 & 36.492 & 40.281 & 52.098 & 16.373 & 37.017 & 15.441 & 25.093 & 25.234 & 44.897 & 15.389 & 27.018 \\
    \textbf{COLD} \cite{he2017writer} & 11.893 & 27.809 & 39.839 & 48.879 &  17.132 & 35.500 & 13.202 & 20.865 & 20.052 & 39.581 & 10.035 & 25.284 \\
    \textbf{CC-Pairs} & 13.480 & 27.652 & 30.932 & 48.039 & 19.892 & 30.180 & 12.209 & 24.278 & 27.492 & 41.033 & 20.840 & 45.923 \\
    \textbf{CC-Triplets} \cite{siddiqi2010text} & 15.415 & 34.732 & 34.893 & 51.289 & 19.010 & 31.122 & 12.757 & 25.207 & 28.565 & 44.028 & 19.982 & 49.284 \\
    \midrule
    \textbf{FragNet} \cite{he2020fragnet} & 69.891 & 85.055 & 86.890 & 93.238 & 77.303 & 92.832 & 48.202 & 71.252 & 54.250 & 81.651 & 64.382 & 90.837 \\
    \textbf{GR-RNN} \cite{he2021gr} & 70.235  & 85.724 & 77.242 & 89.559 & 79.541 & 94.466 & 50.605 & 69.366 & 68.798 & 89.387 & 56.852 & 91.015 \\
    \textbf{SEG-WI} \cite{kumar2020segmentation} & 77.308 & 89.952 & 75.190 & 85.192 & 65.227 & 86.692 & 44.065 & 53.251 & 62.588 & 84.524 & 71.897 & 84.240 \\
    \textbf{Siamese-OWI} \cite{kumar2022siamese} & 81.978 & 92.387 & 84.527 & 92.789 & 50.132 & 89.892 & 55.798 & 70.787 & 70.673 & 86.398 & 72.190 & 93.265 \\
    \textbf{Deep-HWI} \cite{javidi2020deep} & 79.720 & 93.516 & 79.254 & 90.865 & 58.527 & 90.011 & \textbf{60.522} & 71.232 & 65.386 & 90.524 & 54.350 & 75.085 \\ 
    \textbf{SWIS} \cite{manna2022swis} & 64.587 & 80.890 & 86.892 & 94.110 & 49.897 & 82.524 & 40.522 & 62.252 & 56.832 & 73.812 & 55.132 & 76.798 \\ 
    \textbf{SURDS} \cite{chattopadhyay2022surds} & 60.541 & 84.205 & 73.192 & 89.182 & 67.027 & 86.821 & 44.425 & 58.633 & 50.562 & 78.659 & 67.852 & 92.636\\ 
    \midrule
    \textbf{CSSL-RHA (Ours)} & \textbf{83.290} & \textbf{96.101} & \textbf{87.653} & \textbf{95.019} & \textbf{81.004} & \textbf{95.129} & 60.437 & \textbf{76.748} & \textbf{71.178} & \textbf{91.908} & \textbf{82.782} & \textbf{94.355} \\
    \bottomrule
  \end{tabular}
\end{table*}

\begin{table*}
  \centering
  \caption{Accuracy(\%) on EN-HA with different ratios of defects and fake samples. The "\textbf{+}" in the header indicates the proportion of supplemented defect areas/falsified data on the basis of the original data.}
  \label{tab:robustness}
  \begin{tabular}{l|c|ccc|ccc}
    \toprule
    \multirow{2}{*}{\textbf{Methods}} & \multirow{2}{*}{\textbf{Original}} & \multicolumn{3}{c|}{\textbf{Damaged Data}} & \multicolumn{3}{c}{\textbf{Falsified Data}} \\
    \cline{3-8}
    & &  \textbf{+10\%} & \textbf{+30\%} & \textbf{+50\%} & \textbf{+10\%} & \textbf{+20\%} & \textbf{+30\%}\\
    \midrule
    \textbf{SimCLR} \cite{chen2020simple} & 45.893$\pm $0.156 & 44.581$\pm $0.248 & 42.029$\pm $0.215 & 37.036$\pm $0.228 & 42.205$\pm $0.254 & 36.525$\pm $0.133 & 29.685$\pm $0.209 \\
    \textbf{BYOL} \cite{grill2020bootstrap} & 44.732$\pm $0.221 & 44.023$\pm $0.185 & 42.172$\pm $0.189 & 40.375$\pm $0.098 & 42.420$\pm $0.265 & 38.052$\pm $0.145 & 30.561$\pm $0.135 \\
    \textbf{MOCO} \cite{he2020momentum} & 56.256$\pm $0.265 & 55.892$\pm $0.209 & 54.819$\pm $0.092 & 52.128$\pm $0.138 & 54.502$\pm $0.185 & 51.520$\pm $0.263 & 46.520$\pm $0.165 \\
    \midrule
    \textbf{FragNet} \cite{he2020fragnet} & 64.382$\pm $0.342 & 64.017$\pm $0.163 & 63.362$\pm $0.254 & 63.572$\pm $0.170 & 63.528$\pm $0.102 & 61.302$\pm $0.252 & 54.631$\pm $0.263 \\
    \textbf{GR-RNN} \cite{he2021gr} & 56.852$\pm $0.115 & 55.798$\pm $0.129 & 53.250$\pm $0.205 & 49.583$\pm $0.325 & 53.720$\pm $0.455 & 46.202$\pm $0.253 & 38.586$\pm $0.301 \\
    \textbf{SEG-WI} \cite{kumar2020segmentation} & 71.897$\pm $0.175 & 71.179$\pm $0.158 & 70.010$\pm $0.093 & 70.179$\pm $0.165 & 69.018$\pm $0.932 & 66.782$\pm $0.378 & 59.533$\pm $0.345 \\
    \textbf{Siamese-OWI} \cite{kumar2022siamese} & 72.190$\pm $0.246 & 72.564$\pm $0.572 & 70.522$\pm $0.112 & 67.356$\pm $0.165 & 69.472$\pm $0.189 & 60.052$\pm $0.068 & 55.897$\pm $0.277 \\
    \textbf{Deep-HWI} \cite{javidi2020deep} & 54.350$\pm $0.466 & 53.632$\pm $0.140 & 54.198$\pm $0.382 & 53.128$\pm $0.092 & 50.091$\pm $0.163 & 46.455$\pm $0.263 & 41.653$\pm $0.155 \\ 
    \textbf{SWIS} \cite{manna2022swis} & 55.132$\pm $0.215 & 54.289$\pm $0.184 & 52.129$\pm $0.098 & 49.718$\pm $0.136 & 50.893$\pm $0.109 & 43.685$\pm $0.253 & 31.688$\pm $0.363 \\ 
    \textbf{SURDS} \cite{chattopadhyay2022surds} & 67.852$\pm $0.397 & 67.189$\pm $0.713 & 66.892$\pm $0.149 & 64.129$\pm $0.247 & 66.129$\pm $0.148 & 67.137$\pm $0.099 & 65.127$\pm $0.284 \\
    \midrule
    \textbf{CSSL-RHA (Ours)} & \textbf{82.782$\pm $0.332} & \textbf{82.893$\pm $0.248} & \textbf{81.923$\pm $0.115} & \textbf{81.072$\pm $0.038} & \textbf{82.289$\pm $0.429} & \textbf{81.337$\pm $0.042} & \textbf{79.452$\pm $0.169} \\
    \bottomrule
  \end{tabular}
\end{table*}

\subsection{Baseline Models}
\label{Baseline Models}

\textbf{NN-LBP/LPQ/LTP} \cite{chahi2019effective} are three frameworks using different local textural features to characterize the writing styles (i.e., Local Binary Patterns, Local Phase Quantization, and Local Ternary Patterns).

\textbf{CoHinge/QuadHinge} \cite{he2017beyond} are two newly proposed handwriting authentication methods using joint feature distribution (JFD) principles based on the original Hinge kernel.

\textbf{COLD} \cite{he2017writer} is a method use the cloud of line distribution features for writer identification.

\textbf{Chain Code Pairs/Triplets (CC-Pairs/Triplets)} \cite{siddiqi2010text} are automatic writer recognition methods using unconstrained handwritten text images. The only difference is that the number of detection attributes used is 2 and 3 respectively.

\textbf{FragNet} \cite{he2020fragnet} is a recently developed benchmark study for writer identification based on word or text block images that typically contain one word.

\textbf{GR-RNN} \cite{he2021gr} is an end-to-end neural network that can identify writers using handwritten word images. It integrates global-context information with a sequence of local fragment-based features.

\textbf{SEG-WI} \cite{kumar2020segmentation} is a segmentation-free model for writer identification based on a convolution neural network and a weakly supervised region selection mechanism.

\textbf{Siamese-OWI} \cite{kumar2022siamese}  is a novel approach for identifying the writer of a document using input word images. It is text-independent and does not impose any constraints on the size of the input image under examination.

\textbf{Deep-HWI} \cite{javidi2020deep} is an end-to-end handwriting authentication system that relies on a well-designed deep network and efficient feature extraction techniques.

\textbf{SWIS} \cite{manna2022swis} is a self-supervised learning framework for writer independent offline signature verification and expanded in this study for more data.

\textbf{SURDS} \cite{chattopadhyay2022surds} is a two-stage deep learning framework that leverages self-supervised representation learning for writer-independent verification.

\textbf{Self-Supervised Baselines}. We select classic SSL baselines to evaluate the advantages of CSSL-RHA over other SSL frameworks in handwriting authentication (i.e., SimCLR \cite{chen2020simple}, BYOL \cite{grill2020bootstrap}, Barlow Twins \cite{zbontar2021barlow}, and MOCO \cite{he2020momentum}).

\subsection{Results Analysis}
\label{Results Analysis and Comparison}
In order to obtain a more comprehensive and objective evaluation, we conduct extensive experiments on CSSL-RHA from multiple perspectives in this section, including effectiveness, robustness, and interpretability. We futher analyze the results and summarize future work.

\subsubsection{Comparison}
\label{Comparison}
\
\newline
To objectively evaluate the effectiveness of CSSL-RHA, we conduct extensive comparative experiments with baseline models mentioned in Subsection \ref{Baseline Models} on five benchmark datasets and our EN-HA dataset mentioned in Subsection \ref{Datasets}. 

The experimental results are reported in Table \ref{tab:comparison}. The key finding is that our CSSL-RHA algorithm achieves excellent handwriting authentication performance with an average 1.94\% improvement on Top 1 results and more than 2\% improvement on Top 5 results. This breakthrough is accomplished without annotated data and observed in almost every dataset, especially in EN-HA, which simulates real-life data forgery and corruption. We speculate that this is because CSSL-RHA does an excellent job of removing noise while focusing on the most important parts of the data. Other observations from Table \ref{tab:comparison} include: 

\textbf{From features}: Deep learned features provide better results (e.g., FragNet, Deep-HWI, and SEG-WI) than the best handcrafted features (e.g., NN-LBP, NN-LPQ, and NN-LTP). The encoder based on automatic features (e.g., GR-RNN and SEG-WI) has advantages over local texture features (e.g., CoHinge and QuadHinge), and considering structural features (e.g., COLD) even have a negative impact.

\textbf{From datasets}: Most models cannot avoid the negative impact of image forgery, and the Top 1 results of baseline models on EN-HA are far lower than the result of CSSL-RHA, which specifically considers fake samples. Considering the narrowed gap in the Top 5 results, it indicates that balancing samples and finding key areas may be the key to improving performance.

\textbf{From learning paradigm}: Compared with other SSL frameworks, our CSSL-RHA achieves great improvement, illustrating the complexity and specificity of handwriting features require specific settings to learn.


\subsubsection{Robustness Study}
\label{Robustness Study}
\
\newline
To validate the robustness of CSSL-RHA, we conduct comparative experiments based on fake data and flawed data, adopting the EN-HA dataset specially constructed for real-world handwriting authentication applications. We set and adjust the ratio of defect images and fake images to ideal samples in EN-HA, and record the average results of ten experiments. Based on the results in Table \ref{tab:comparison}, we select ten baselines with better performance for comparison. 

Table \ref{tab:robustness} shows the accuracy test results of each model under non-ideal data. The results demonstrate that CSSL-RHA has an advantage in robustness, achieving similar handwriting authentication performance in datasets containing more noise than in ideal data. Specifically, when the proportion of falsified data increases by 10\%, the performance of CSSL-RHA hardly decreases (unlike other frameworks, which decrease by an average of 6\%). When the proportion of falsified data increases to 30\%, this advantage expands to more than 15\%. In the case of damaged data (with stains, scratches, etc.), when the noise ratio reaches 50\%, the performance of CSSL-RHA only drops by 1\%, exceeding other algorithms. These findings validate the robustness of CSSL-RHA and its advantages in practical applications, especially in historical digital archives where the damage of ancient manuscripts due to improper storage and transportation is typically more than 30\%.

\begin{table*}
  \caption{Ablation study of CSSL-RHA on EN-HA. The "$\checkmark$" indicates that the corresponding module is activated in this round of testing. The "D-N\%" and "F-N\%" after "Accuracy(\%)" represent the area ratio of noise and the ratio of falsified samples increased in the EN-HA participating in the experiment, respectively.}
  \label{tab:ablation}
  \begin{tabular}{l|l|ccccccc}
    \toprule
    \multirow{4}{*}{Encoders} & ViT-Base     &$\checkmark$&$\checkmark$&$\checkmark$&$\checkmark$&  &  &  \\
    & ViT-Tiny &  &  &  &  & $\checkmark$ &  &  \\
    & ViT-Small &  &  &  &  &  &$\checkmark$ &  \\
    & Swin &  &  &  &  &  &  &$\checkmark$ \\
    \cline{1-2}   
    \multicolumn{2}{l|}{Information-theoretic Filter} &  &$\checkmark$ & &$\checkmark$&$\checkmark$&$\checkmark$&$\checkmark$\\   
    \multicolumn{2}{l|}{Adaptive Matching} &  &  & $\checkmark$ &$\checkmark$ &$\checkmark$&$\checkmark$&$\checkmark$ \\
    \midrule    
    \multicolumn{2}{l|}{\textbf{Accuracy(\%)}} & 76.429$\pm$0.284 & 79.957$\pm$0.247 & 81.447$\pm$0.148 & 82.782$\pm$0.354 & 80.578$\pm$0.323 & 80.909$\pm$0.693 & 83.468$\pm$0.150  \\
    \multicolumn{2}{l|}{\textbf{Accuracy(\%)-D-10\%}} & 75.544$\pm$0.182 & 79.462$\pm$0.190 & 81.090$\pm$0.138 & 82.893$\pm$0.248 & 80.542$\pm$0.155 & 82.655$\pm$0.147 & 83.893$\pm$0.120 \\
    \multicolumn{2}{l|}{\textbf{Accuracy(\%)-D-30\%}} & 73.027$\pm$0.099 & 77.372$\pm$0.127 & 79.452$\pm$0.173 & 81.923$\pm$0.115 & 79.849$\pm$0.218 & 80.128$\pm$0.174 & 81.892$\pm$0.378 \\
    \multicolumn{2}{l|}{\textbf{Accuracy(\%)-D-50\%}} & 70.572$\pm$0.245 & 76.289$\pm$0.138 & 77.541$\pm$0.156 & 81.072$\pm$0.038 & 79.218$\pm$0.092 & 80.178$\pm$0.77 & 83.259$\pm$0.100 \\
    \multicolumn{2}{l|}{\textbf{Accuracy(\%)-F-10\%}} & 73.789$\pm$0.238 & 73.027$\pm$0.098 & 81.561$\pm$0.189 & 82.289$\pm$0.329 & 81.218$\pm$0.135 & 81.355$\pm$0.163 & 83.189$\pm$0.274 \\
    \multicolumn{2}{l|}{\textbf{Accuracy(\%)-F-20\%}} & 70.947$\pm$0.177 & 70.204$\pm$0.155 & 80.848$\pm$0.151 & 81.337$\pm$0.042 & 79.544$\pm$0.531 & 81.863$\pm$0.150 & 80.520$\pm$0.099 \\
    \multicolumn{2}{l|}{\textbf{Accuracy(\%)-F-30\%}} & 65.263$\pm$0.210 & 66.240$\pm$0.085 & 78.165$\pm$0.377 & 79.452$\pm$0.169 & 76.237$\pm$0.098 & 77.897$\pm$0.179 & 78.855$\pm$0.166 \\
    \bottomrule
  \end{tabular}
\end{table*}

\subsubsection{Ablation Study}
\label{Ablation Study}
\
\newline
To examine the interpretability of our proposed CSSL-RHA, we conduct comprehensive ablation studies in this section. We build CSSL-RHA modularly to evaluate the combined effect of different encoders, an information-theoretic filter, and the proposed adaptive matching scheme on the improvement of the performance. We also set different degrees of abnormal samples for further exploration.

Table \ref{tab:ablation} demonstrates the effectiveness of each module in terms of stability. From the results, we can observe that our proposed filter and the adaptive matching scheme are effective in improving the accuracy, achieving about 3\% and 7\% improvements on EN-HA, respectively. Comparing different degrees of anomalies, it can be found that these increases are more pronounced, that is, slowing down the negative impact of abnormal samples when there are more damaged or falsified data. For encoders, we use two structures, ViT and Swin, to calculate the effect of different patch segmentation and representation strategies. Among them, we consider three variants of ViT, namely ViT-Tiny, ViT-Small, and ViT-Base, with different embedding sizes as described in \cite{yang2022reading}. The results indicate that the hierarchical Transformers structure is helpful in eliminating redundant patches, and ViT-Base with a larger embedding size achieves higher detection accuracy.

\subsubsection{Further Analysis and Future Work}
\label{Further analysis and Limitations}
\
\newline
CSSL-RHA has shown superior results in handwriting authentication even under conditions with a lot of interference. However, despite the above breakthroughs, handwriting authentication remains an open area for research due to the various challenges it presents, which we aim to explore further in the future.

Firstly, analyzing the structural information of unstructured handwriting features is a difficult but valuable task for representation learning. Our proposed CSSL-RHA is not content-aware and mainly relies on fine-grained features to make determinations, without considering the aforementioned issue. Secondly, changes in handwriting may occur due to various factors such as mood, mental state, age, and situation. Additionally, different writing instruments can also lead to different handwriting samples. Therefore, we plan to introduce the temporal dimension into the classification and feature learning of handwriting, rather than limiting it to the identification of individual style fixations. Furthermore, supervised methods for handwriting authentication are task-specific due to the different writing properties of different languages (such as the cursive form of Arabic). Our CSSL-RHA is suitable for such complex and challenging changes, but the lack of evaluation standards and benchmark datasets hinders its development. We believe this work can provide a solid foundation for handwriting authentication and will further expand our research from the perspectives mentioned above in the future.


\section{Conclusion}
\label{Conclusion}
In this paper, we propose a novel framework, named Contrastive Self-Supervised Learning for Robust Handwriting Authentication (CSSL-RHA), to dynamically learn the complex handwriting features for writer identity detection task without supervision. It is composed of four stages which includes several innovations: pre-processing, generalized pre-training, personalized calibration, and personal testing. Specifically, in the pre-processing stage, we develop an information-theoretic filter to remove noise from defects such as scratches and stains. In the pre-training stage, we propose an adaptive matching scheme to reweight images into patches representing local regions likely to be dominated by more important elements, which helps minimize the negative impact of falsified and damaged data. To establish semantic correspondences between patches, we present contrastive self-supervised training with a momentum-based paradigm specific to handwriting authentication, entirely without relying on annotations. The last two stages efficiently validate the pre-training results and verify the identity of unknown handwriting samples based on the trained model. Meanwhile, we collect 800 English manuscripts from 20 famous historical figures and 20 volunteers to embrace more challenging scenarios for handwriting authentication in reality. Extensive experiments demonstrate that our proposed method outperforms existing baselines, highlighting the effectiveness and robustness of CSSL-RHA for the task of handwriting authentication.


\bibliographystyle{ACM-Reference-Format}
\bibliography{References}

\end{document}